\documentclass[12pt,a4paper]{article}
\usepackage[utf8]{inputenc}
\usepackage[english]{babel}
\usepackage{amsmath}
\usepackage{amsfonts}
\usepackage{amssymb}
\usepackage{float}
\usepackage{graphicx}
\usepackage[colorlinks=true, citecolor=magenta, linkcolor=magenta]{hyperref}
\usepackage{xcolor}
\usepackage[left=1in,right=1in,top=1in,bottom=1in]{geometry}
\usepackage{subcaption}
\usepackage{natbib}

\usepackage{graphicx}
\usepackage{caption}
\usepackage{subcaption}
\usepackage{float} 
\usepackage{multirow}
\usepackage{booktabs}
\usepackage{makecell}
\usepackage{adjustbox}

\usepackage{authblk}
\title{AKReF: An argumentative knowledge representation framework for structured argumentation}
\author[ ]{Debarati Bhattacharjee}
\author[ ]{Ashish Anand}

\affil[ ]{Department of Computer Science and Enginering,}
\affil[ ]{IIT Guwahati, Assam, India, 781039}
\affil[ ]{\{b.debarati, anand.ashish\}@iitg.ac.in}

\date{}  
\begin{document}
\maketitle

\begin{abstract}
\noindent This paper presents a framework to convert argumentative texts into argument knowledge graphs (AKG). The proposed argumentative knowledge representation framework (AKReF) extends the theoretical foundation and enables the AKG to provide a graphical view of the argumentative structure that is easier to understand. Starting with basic annotations of argumentative components (ACs) and argumentative relations (ARs), we enrich the information by constructing a knowledge base (KB) graph with metadata attributes for nodes. Next, we apply modus ponens on premises and inference rules from the KB to form arguments. From these arguments, we create an AKG. The nodes and edges of the AKG have attributes that capture important argumentative features such as the type of premise (e.g., axiom, ordinary premise, assumption), the type of inference rule (e.g., strict, defeasible), the preference order over defeasible inference rules, markers (e.g., ``therefore", ``however" etc.), and the type of attack (e.g., undercut, rebuttal, undermining), among others. We identify inference rules by locating a specific set of markers, called inference markers (IM). This, in turn, makes it possible to identify undercut attacks previously undetectable in existing datasets. AKG prepares the ground for reasoning tasks, including checking the coherence of arguments and identifying opportunities for revision. For this, it is essential to find indirect relations, many of which are implicit. Our proposed AKG format, with annotated inference rules and modus ponens, will help reasoning models learn the implicit, indirect relations that require inference over arguments and the relations among them. We use an essay from the AAEC dataset to illustrate the proposed framework for constructing an AKG. We further show the application of AKG in complex analyses such as extraction of conflict-free set of arguments and a maximal set of admissible arguments.
\end{abstract}

\section*{Keywords}
argument mining, argumentation framework, marker, argument knowledge graph, knowledge base, graph-based reasoning


\section{Introduction}
Argument mining (AM) \cite{lawrence2020argument} is a key area of natural language processing (NLP) that aims to identify and understand arguments in text or speech. Most existing work in AM focuses on identifying argumentative components (ACs) \cite{feng2011classifying} and their associated argumentative relations (ARs) \cite{carstens2015towards}. However, a deeper goal of AM, especially in downstream reasoning tasks, is to assess the structure, coherence, and dynamics of argumentation — including the potential need for belief revision \cite{snaith2017argument}. This requires inference over the full set of arguments and the relations between them. Many of these relations are not directly stated and must be inferred through intermediate or implicit argumentative links \cite{stab2014identifying}. Existing methods typically capture only direct relations. Identifying such indirect inferences, most of which are implicit, remains a significant challenge in the field.

Theoretical argumentation began with Dung's abstract argumentation framework \cite{dung1995acceptability}. In this model, arguments are abstract objects. The only relation between them is attack. Semantics like complete, preferred, stable, and grounded define acceptable argument sets. Later, researchers added more details to the arguments. For example, the ASPIC framework \cite{amgoud2006final} instantiated Dung’s abstract argumentation framework with structured arguments. The subsequent \text{ASPIC}$^{+}$ framework \cite{prakken2010abstract} introduced undermining attacks and differentiated among four types of premises. Attack relations were partly resolved using preferences.
In argumentation systems, even small changes to the knowledge base (KB) can cause effects that are problematic to analyze. Snaith and Reed \cite{snaith2017argument} addressed this problem using an argument revision model that identified minimal, intuitive, and justifiable changes. While these frameworks are powerful, manually analyzing large argumentative texts is labor-intensive due to their complex theoretical structures and flat representations. The identification and analysis of indirect and implicit inferences remain challenging. This work aims to provide a systematic approach for converting ACs and ARs into an argumentation framework represented as a heterogeneous graph (HG). This graphical representation is relatively easier to interpret compared to textual formats, which often require domain expertise. Furthermore, such a representation facilitates training models for logical reasoning tasks.

We propose a framework named the argumentative knowledge representation framework (AKReF) to formally represent a KB graph and derive an argument knowledge graph (AKG) from it. The KB graph includes \textit{premises} and \textit{inference rule premises} as nodes, and their interrelations (\textit{contrary} and \textit{agreement}) as edges. We identify inference rules by locating a specific set of markers, which we term inference markers (IM). Inference rules are included in the KB as a special type of \textit{premise}, called the \textit{inference rule premise}. Each node in the KB has its own metadata as attributes, while the edges do not. Subsequently, attributes are incorporated into the  AKG to add additional information that is missing in the basic annotation. This additional information is essential in successive steps for belief revision \cite{snaith2017argument} and argument evaluation \cite{baroni2015automatic, wachsmuth2017argumentation}. The proposed KB expands automatically when a new component or relation is introduced. This expandability makes it suitable for real-world analysis, especially for KBs with increasing complexity. We then derive a set of arguments from the graphical representation of the KB. The AKG is constructed from this argument set. It includes three types of nodes: \textit{premises} (P), \textit{inference rule premises} (IRP), and \textit{conclusions} (C). The edges represent their interrelations — \textit{support}, \textit{attack}, and \textit{modus ponens}. Each node and \textit{attack} edge in the AKG has its own attributes. \textit{Support} and \textit{modus ponens} edges do not have attributes. We integrate all three types of attacks — undercutting, rebuttal, and undermining \cite{prakken2010abstract} 

\pagebreak
\begin{figure}[H]
    \centering
    \includegraphics[width=0.95\linewidth, height=\textheight, keepaspectratio]{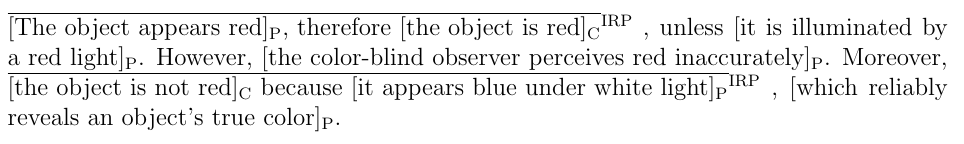}
    \caption{Pollock’s classic example of undercutting \cite{pollock1987defeasible} is extended, and annotations are incorporated. P, C, and IRP denote \textit{premise}, \textit{conclusion}, and \textit{inference rule premise}, respectively. Text spans annotated as P or C are enclosed in square brackets. Spans annotated as IRP are marked using an overline.}
    \label{fig:image1}
\end{figure}

\begin{figure}[htbp]
  \centering
  \begin{subfigure}[b]{0.45\textwidth}
    \centering
    \includegraphics[width=\textwidth]{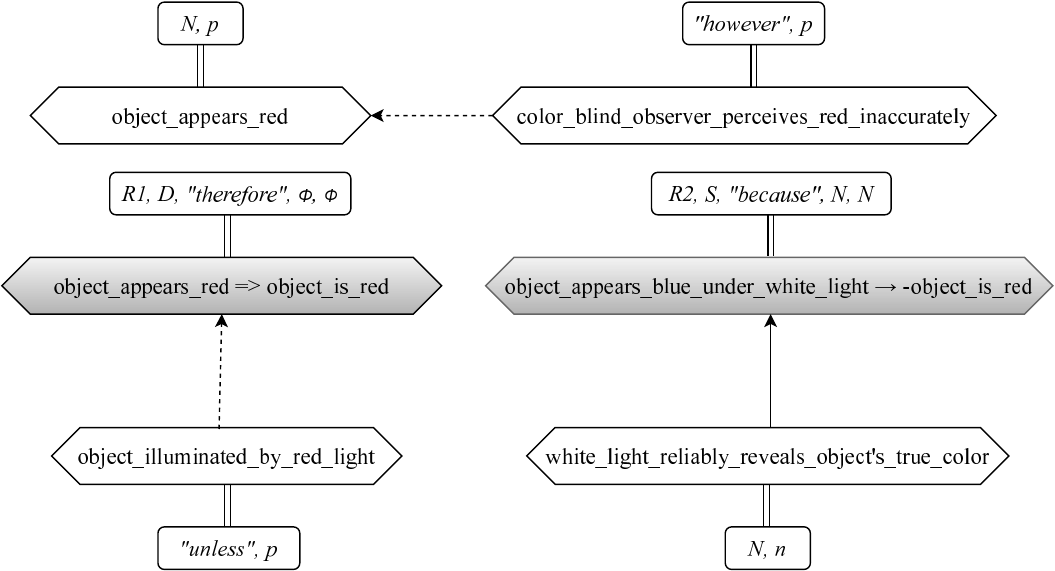}
    \caption{The KB Graph}
    \label{fig:image2}
  \end{subfigure}
  \hspace{1mm}
  \begin{subfigure}[b]{0.5\textwidth}
    \centering
    \includegraphics[width=\textwidth]{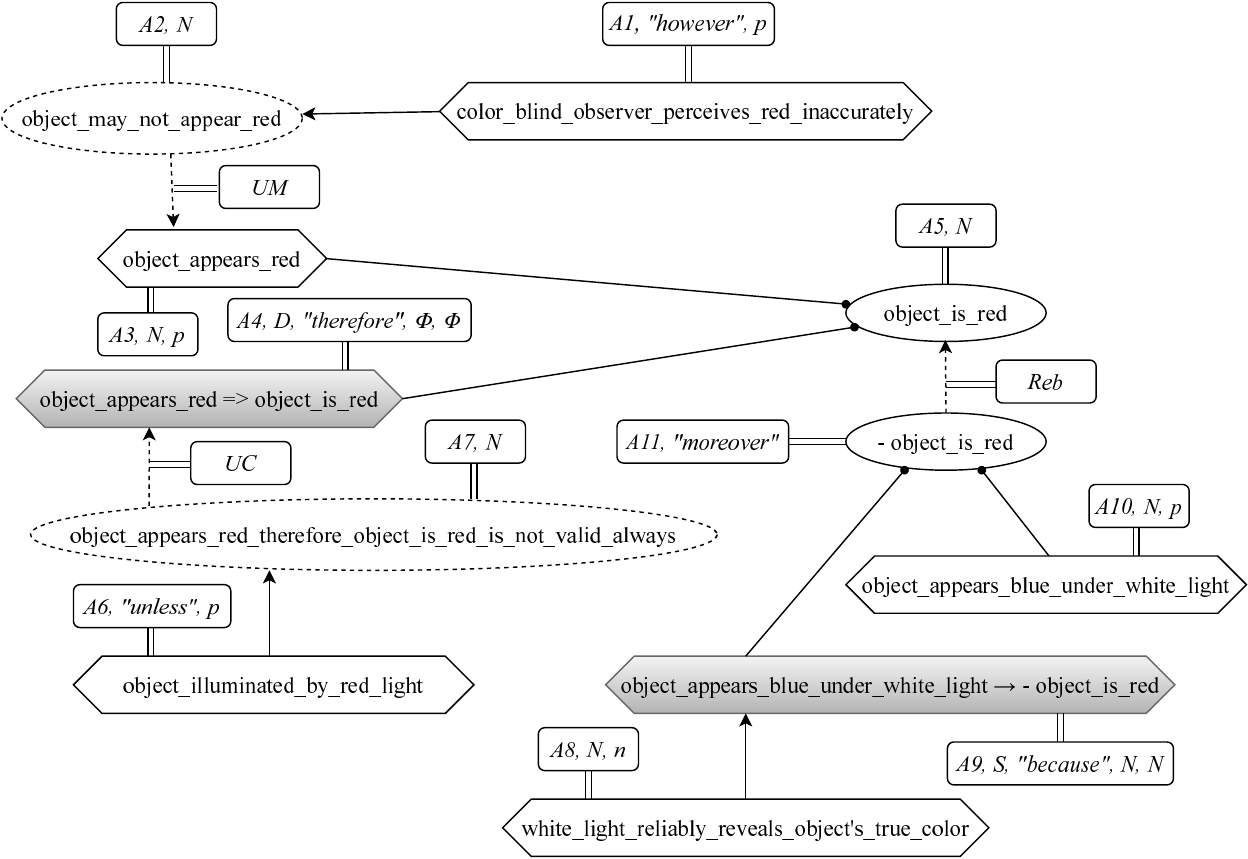}
    \caption{The AKG}
    \label{fig:image3}
  \end{subfigure}
  \caption{Illustration of the KB Graph and the AKG for the constructed example in Figure \ref{fig:image1}. Refer to Figure \ref{fig:image4} for the details of the symbols used for different types of nodes, edges, and attributes. The detailed methodology for the creation of the KB graph and the AKG is provided in Sections 3.1 and 3.2.}
  \label{fig:combined}
\end{figure}

\noindent — into the AKG. Additionally, we represent \textit{implicit premises} and \textit{implicit claims} in enthymemes using dotted nodes in the KB graph and the AKG. Details of the attributes in the KB graph and AKG are provided in Sections 3.1 and 3.2, respectively. In the following, we build upon Pollock’s classic example of undercutting \cite{pollock1987defeasible} (\textit{The object appears red therefore the object is red unless it is illuminated by a red light.}), as shown in Figure~\ref{fig:image1}, incorporating annotation to illustrate the proposed KB graph and AKG structure. The corresponding KB graph and AKG, emerging from our method, are shown in Figure \ref{fig:combined}. The detailed step-by-step method for constructing a KB graph and an AKG from an existing dataset is discussed in Sections 3.1 and 3.2.

This paper makes five main contributions. (1) We enrich basic dataset annotations by converting them into a KB graph and add metadata as attributes to the nodes. (2) The premises and inference rules in the KB graph are used to construct arguments through modus ponens. (3) We construct the novel AKG from the argument set. It has attributes on its nodes and edges. (4) Modus ponens edges are introduced in the AKG to identify indirect relations. (5) Existing annotations do not capture inference rules. We identify both strict and defeasible rules using IMs, a set of markers for this purpose. We also add the inference rules as a new type of premise node in both the KB and the AKG. This helps us to detect undercut attacks, which were not possible to identify in the existing datasets.

\section{Background}
This section reviews key definitions from the literature, with a focus on argument structures, attack relations, inference mechanisms, and acceptability semantics. Although these definitions exist in previous works, we include them here for the sake of completeness. In the works of Dung \cite{dung1995acceptability}, arguments are considered as abstract entities upon which attack relation are established. Based on these arguments and the relations among them, argumentation framework is defined.

\begin{enumerate}

\item \textbf{Argumentation Framework:} An argumentation framework (AF) is defined as a pair $AF = (Args, Atts)$, where $Args$ represents a collection of arguments, and $Atts \subseteq Args \times Args$ denotes a binary relation capturing the attacks between arguments \cite{dung1995acceptability}.

\hspace*{2em}The ASPIC\textsuperscript{+} framework \cite{prakken2010abstract} offers a structured realization of Dung's model. Its key components consist of an argumentation system along with a knowledge base.

\item \textbf{Argumentation System:} An argumentation system (AS) is represented as a tuple $AS = (L, \overline{\phantom{x}}, R, \leq)$, where $L$ denotes a logical language; $\overline{\phantom{x}}$ is a contrariness function mapping elements from $L$ to subsets of $L$, i.e., $2^L$; $R = R_s \cup R_d$ consists of strict inference rules ($R_s$) and defeasible inference rules ($R_d$), with no overlap between them ($R_s \cap R_d = \emptyset$); and $\leq$ defines a partial pre-order relation on the set of defeasible rules $R_d$ \cite{prakken2010abstract}.

\begin{enumerate}
\item \textbf{Logical Language:} Consider $L$ as a logical language, and let $\overline{\phantom{x}}$ be a contrariness function that maps elements from $L$ to subsets of $L$, i.e., $2^L$. If $\phi \in \overline{\psi}$ but $\psi \notin \overline{\phi}$, then $\phi$ is referred to as a contrary of $\psi$. In contrast, if both $\phi \in \overline{\psi}$ and $\psi \in \overline{\phi}$ hold, then $\phi$ and $\psi$ are said to be contradictory, denoted as $\phi = - \psi$ \cite{prakken2010abstract}.

\item \textbf{Inference Rules:} Inference rules are of two types, as follows:
\begin{itemize}
\item A \textbf{strict inference rule} has the structure $\phi_1, \ldots, \phi_n \rightarrow \phi$, which informally signifies that whenever $\phi_1, \ldots, \phi_n$ are true, $\phi$ must also be true \textit{in all cases without exception} \cite{prakken2010abstract}.

\item A \textbf{defeasible inference rule} takes the form $\phi_1, \ldots, \phi_n \Rightarrow \phi$, which informally suggests that when $ \phi_1, \ldots, \phi_n$ are true, it is \textit{presumed} that $\phi$ holds \cite{prakken2010abstract}.
\end{itemize}

The terms $\phi_1, \ldots, \phi_n$ are called the \textit{antecedents} of the rule, whereas $\phi$ represents its \textit{consequent}.
\end{enumerate}

\item \textbf{Knowledge Base:} Within an AS $(L, \overline{\phantom{x}}, R, \ \leq)$, a knowledge base (KB) is defined as a pair $(K,\leq')$, where $K \subseteq L$ and can be partitioned as $K = K_n \cup K_p \cup K_a$, satisfying $K_n \cap K_p \cap K_a = \emptyset$. The components are characterized as follows:
\begin{itemize}
\item $K_n$ contains the (necessary) \textit{axioms}. Premises classified as axioms are immune to attacks;
\item $K_p$ includes the \textit{ordinary premises}. These premises may be targeted by attacks, where the outcome is influenced by the preference ordering among arguments;
\item $K_a$ comprises the \textit{assumptions}. Assumption premises are always vulnerable, with any attack against them being automatically successful.
\end{itemize}
The relation $\leq'$ represents a partial pre-order defined over $K \setminus K_n$ \cite{snaith2017argument}.

\item \textbf{Attack:} In the context of an argument, the function $Prem$ extracts all formulas from $K$ (referred to as premises) that participate in constructing the argument. The function $Conc$ yields the conclusion of the argument, whereas $Sub$ retrieves all its subarguments \cite{prakken2010abstract}. There are three types of possible attacks on an argument:

\begin{itemize}
\item \textbf{Undermining:} An argument may be subjected to an \textbf{attack on a non-axiom premise}, known as undermining. Argument $A$ undermines $B$ (on $\phi$) if $Conc(A) \in \overline{\phi}$ for some $\phi \in Prem(B) \setminus K_n$. Furthermore, $A$ is said to contrary-undermine $B$ if $Conc(A)$ is a contrary of $\phi$ or if $\phi \in K_a$ \cite{prakken2010abstract}.

\item \textbf{Rebutting:} This form of attack targets the \textbf{conclusion} of an argument. Argument $A$ rebuts argument $B$ (on $B'$) if $Conc(A) \in \overline{\phi}$ for some $B' \in Sub(B)$ where $B'$ takes the form $B''_1, ..., B''_n \Rightarrow \phi$ \cite{prakken2010abstract}.

\item \textbf{Undercutting:} In this attack, the \textbf{defeasible inference rule} applied in the argument is challenged. Argument $A$ undercuts $B$ (on $B'$) if $Conc(A) \in \overline{B'}$ for some $B' \in Sub(B)$ expressed as $B''_1, ..., B''_n \Rightarrow \psi$ \cite{prakken2010abstract}.
\end{itemize}

\item \textbf{Conflict-free:} A set \textit{S} of arguments is considered conflict-free when no pair of arguments \textit{A} and \textit{B} within \textit{S} exists such that \textit{A} attacks \textit{B} \cite{dung1995acceptability}.

\item \textbf{Acceptability:} A set $S$ of arguments is said to attack an argument $B$ (or equivalently, $B$ is attacked by $S$) if there exists some argument within $S$ that attacks $B$ \cite{dung1995acceptability}.\\
\hspace*{2em}An argument $A \in Args$ is deemed acceptable relative to a set \textit{S} of arguments if, for every argument $B \in Args$, whenever \textit{B} attacks \textit{A}, it holds that \textit{B} is attacked by some member of \textit{S} \cite{dung1995acceptability}.

\item \textbf{Admissibility:} A set of arguments \textit{S} is admissible if it is conflict-free and every argument in \textit{S} is acceptable with respect to \textit{S} \cite{dung1995acceptability}.

\item \textbf{Naïve Semantics:} In a given AF, naïve semantics refer to conflict-free sets of arguments that are maximal with respect to set inclusion \cite{ruiz2022automatic}.

\item \textbf{Preferred Semantics:} In a given AF, preferred semantics are characterized as admissible sets of arguments that are maximal with respect to set inclusion \cite{ruiz2022automatic}.
\end{enumerate}

\section{Methodology}
This section presents the AKReF, describing the process of constructing the KB graph and, subsequently, the AKG, starting with the essential definitions. We build upon existing concepts in the literature to incorporate them into our method.

\begin{enumerate}
\item \textbf{The Set of Contraries in a KB:} The set of all contraries in a KB is denoted by $C(KB)$. Formally, $C(KB) = \bigcup_{\phi \in KB} \overline{\phi}$, where $\overline{\phi}$ represents the set of contraries of the formula $\phi \in L$, where $L$ denotes a logical language.

\pagebreak
\begin{figure}[H] 
    \centering
    \includegraphics[width=14cm, height=6cm]{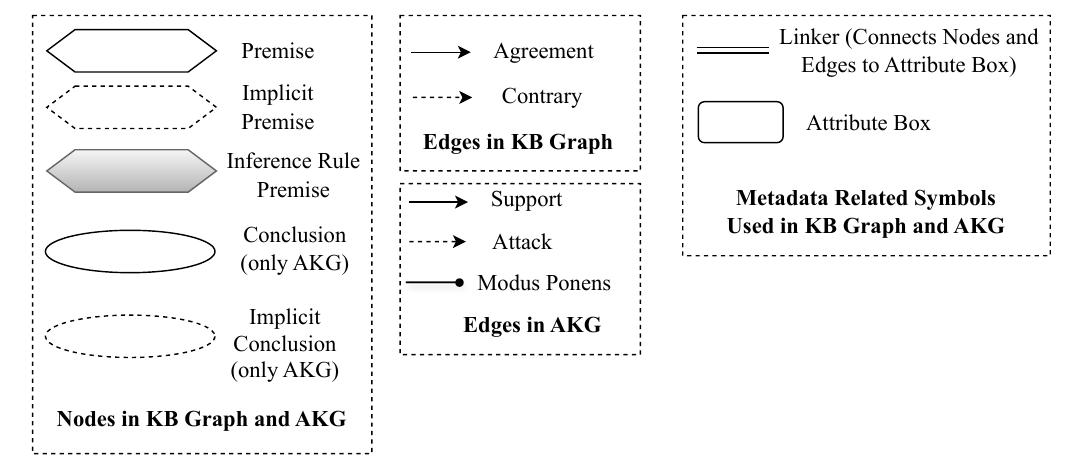}
    \caption{Symbols used in KB Graph and AKG}
    \label{fig:image4}
\end{figure}

\item \textbf{Agreement Function and the Set of Agreements in a KB:} We propose an agreement function, similar to the contrariness function available in the literature and discussed in Section 2.\\
\hspace*{2em}Let $L$ be a set representing a logical language, and let $\underline{\phantom{x}}$ be an agreement function mapping elements of $L$ to $2^L$. When $\phi \in \underline{\psi}$, it indicates that $\phi$ \textit{agrees with} $\psi$. If both $\phi \in \underline{\psi}$ and $\psi \in \underline{\phi}$ hold, meaning they \textit{mutually agree}, we express this by writing $\phi = \psi$. In this case, $\phi$ and $\psi$ exhibit symmetric agreement.\\
\hspace*{2em}Let $Ag(KB)$ denote the set of all agreements in a KB.
Then, $Ag(KB) = \bigcup_{\phi \in KB} \underline{\phi}$, where $\underline{\phi}$ represents the set of all agreements of a formula $\phi \in L$.

\item \textbf{Extended Argumentation System:} We extend the AS discussed in Section 2. An extended AS (EAS) is defined as a tuple $EAS = (L, \overline{\phantom{x}}, \underline{\phantom{x}}, R, \leq)$, where $L$ represents a logical language; $\overline{\phantom{x}}$ denotes a contrariness function, and $\underline{\phantom{x}}$ denotes an 
agreement function, both mapping elements of $L$ to $2^L$; $R = R_s \cup R_d$ denotes the set of inference rules, consisting of strict rules $R_s$ and defeasible rules $R_d$, with $R_s \cap R_d = \emptyset$; and $\leq$ is a partial pre-order defined over $R_d$.

\item \textbf{Extended Knowledge Base:} We extend the KB discussed in Section 2. Within an EAS $(L, \overline{\phantom{x}}, \underline{\phantom{x}}, R , \leq)$, an extended KB (EKB) is defined as a quadruple $(K, C(KB),$\\$ Ag(KB), \leq')$, where $K \subseteq L$ and decomposed as $K = K_n \cup K_p \cup K_a$, with the condition $K_n \cap K_p \cap K_a = \emptyset$. The subsets $K_n$, $K_p$, and $K_a$ are specified in Section 2. The relation $\leq'$ is a partial pre-order on $K \setminus K_n$, while $C(KB)$ and $Ag(KB)$ follow the definitions provided in Section 3.
\end{enumerate}

\subsection{Construction of a KB graph}
Our proposed KB graph is represented as a HG. The KB graph consists of three types of nodes and two types of edges. The three types of nodes represent \textit{premises}, \textit{implicit premises} (for enthymemes), and \textit{inference rule premises}. Two types of edges are dedicated to \textit{agreement} and \textit{contrary} relations. The attributes of a node are presented in a rectangular box, attached to that node with a link. All the components of a KB graph are illustrated in Figure \ref{fig:image4}.

In a KB graph, and subsequently in an AKG, an attribute is a piece of descriptive information attached to a node or an edge that adds semantic information beyond the

\newpage
\begin{table}[H]
\centering
\small
\caption{For components present in both the KB graph and the AKG, separate attributes and values are shown for each graph type. For components appearing only in one graph, the graph type is indicated. Value examples illustrate one or more possible scenarios. For the full range of values, see Sections 3.1 and 3.2.}
\label{tab:table1}
\begin{adjustbox}{max width=\linewidth}
\begin{tabular}{@{}lll@{}}
\toprule
\textbf{Component} & \textbf{Attribute} & \textbf{Value Example} \\
\midrule
\multirow{2}{*}{\texttt{Premise/ Implicit Premise}} 
    & ID (for AKG only)                               & A$_4$\\ 
    & Marker (for both KB graph and AKG)              & for example, for instance, furthermore \\
    & Type (for both KB graph and AKG)                & n (Axiom) / p (Ordinary) / a (Assumption) \\
\midrule
\multirow{2}{*}{\makecell[tl]{\texttt{Conclusion/ Implicit Conclusion}\\\texttt{(Only in AKG)}}}
    & ID              & A$_5$ \\
    & Marker          & clearly, in short, in conclusion \\
\midrule
\multirow{5}{*}{\texttt{Inference Rule Premise}} 
    & ID                               & \makecell[l]{R$_1$ (for KB graph)\\ A$_1$ (for AKG)}\\
    & Type (for both KB graph and AKG) & S (Strict) / D (Defeasible) \\
    & IM     & because, hence, thus \\
    & L$_1$ (Set of less preferred rules) & \makecell[l]{\{R$_2$, R$_3$\}; N, if rule type is S (for KB graph)\\ \{A$_2$, A$_3$\}; N, if rule type is S (for AKG)}\\
    & L$_2$ (Set of more preferred rules) & \makecell[l]{$\phi$; N, if rule type is S (for KB graph)\\ $\phi$; N, if rule type is S (for AKG)}\\
\midrule
\makecell[l]{\texttt{Agreement (Edge)}\\\texttt{(Only in KB graph)}} 
    & \makecell[c]{-} & \makecell[c]{-} \\
\midrule
\makecell[l]{\texttt{Contrary (Edge)}\\\texttt{(Only in KB graph)}} 
    & \makecell[c]{-} & \makecell[c]{-} \\
\midrule  
\makecell[l]{\texttt{Support (Edge)}\\\texttt{(Only in AKG)}} 
    & \makecell[c]{-} & \makecell[c]{-} \\
\midrule
\makecell[l]{\texttt{Attack (Edge)}\\\texttt{(Only in AKG)}} 
    &\makecell[c]{Type}   & Reb (Rebuttal) / UM (Undermine) / UC (Undercut) \\
\midrule
\makecell[l]{\texttt{Modus Ponens (Edge)}\\\texttt{(Only in AKG)}} 
    & \makecell[c]{-} & \makecell[c]{-} \\
\bottomrule
\end{tabular}
\end{adjustbox}
\vspace{0.5em}

\noindent\parbox{\linewidth}{
\footnotesize 
``-" denotes that the component has no attribute.
}
\end{table}

\noindent basic topology of the graph. They encode properties relevant to the argumentative structure, such as type, discourse cue, inference rule category, preference order, etc. The attribute box, which contains the attributes of a node or edge, is connected to the corresponding graph component by a linker. An attribute has two main features: (1) it is attached to a node or edge and is not a separate entity; (2) it describes some property of the node or edge to which it is attached. Table \ref{tab:table1} presents a list of attributes along with one or more example values for each attribute in both the KB graph and the AKG.

The attribute box in the KB graph corresponding to an \textit{inference rule premise} stores various attributes in the following order: the rule ID, the type of the rule ($S$ for strict, $D$ for defeasible), the IM that identifies the inference rule, and the sets $L_1$, and $L_2$. If the rule type is $D$, set $L_1$ contains the IDs of all defeasible rules that have lower preferences than this rule. Similarly, $L_2$ set contains the IDs of all defeasible rules that have greater preferences than this rule. $L_1$ and $L_2$ can be empty sets. If the rule type is $S$, we use $N$ in place of $L_1$ and $L_2$, where $N$ denotes $None$ because the ASPIC\textsuperscript{+} framework does not impose a preference order on strict rules. A potential set of arguments can be $\{R_1, D, ``therefore", \{R_2, R_3\}, \phi\}$, where $R_1$ is the ID of the inference rule. $D$ denotes that it is defeasible. Next, “\textit{therefore}” is the IM associated with the inference rule $R_1$. $R_2$ and $R_3$ have lower preferences than $R_1$. $\phi$ indicates that there \noindent is currently no rule in the KB with a greater preference than $R_1$.

The attributes of a \textit{premise} node include any marker contained in the \textit{premise} and the type of \textit{premise} as mentioned in Section 2. In the absence of a marker, the first field is filled with $N$. For the second field, the letter $n$ denotes \textit{axioms}, $p$ denotes \textit{ordinary}

\pagebreak
\begin{figure}[H] 
    \centering
    \includegraphics[width=0.95\linewidth, height=\textheight, keepaspectratio]{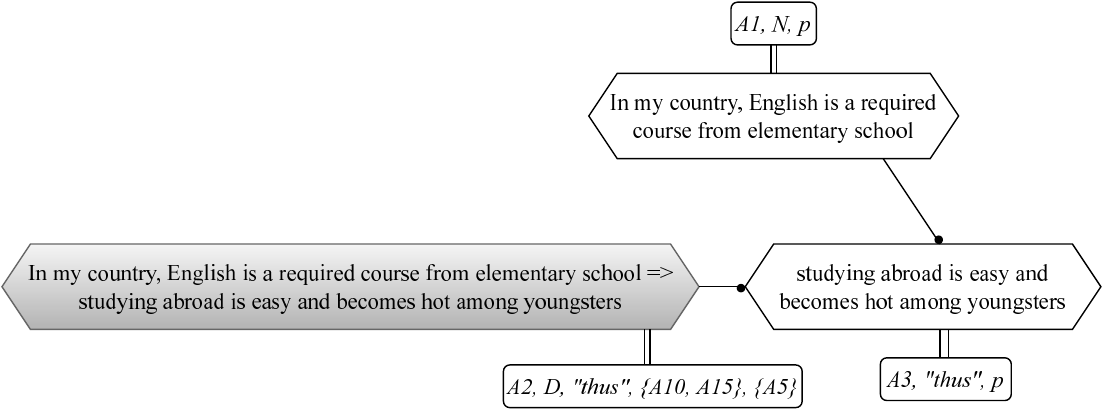}
    \caption{In a portion of an AKG created from \textit{essay056} of the \textit{AAEC version2} dataset \cite{stab2017parsing}, the \textit{modus ponens} edges connect arguments $A_1$ and $A_2$ to argument $A_3$, which is a premise rather than a conclusion.}
    \label{fig:image5}
\end{figure}

\begin{figure}[htbp] 
  \centering
  \begin{subfigure}[b]{0.45\textwidth}
    \centering
    \includegraphics[width=\textwidth]{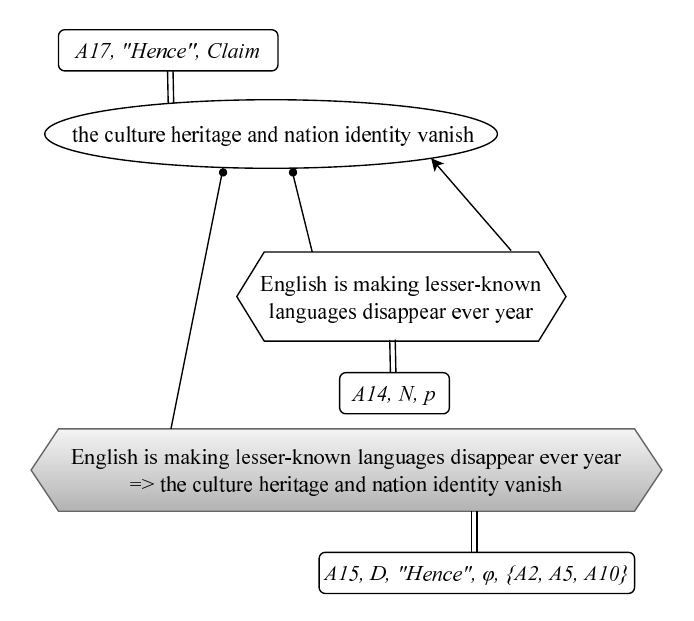}
    \caption{}
    \label{fig:image6}
  \end{subfigure}
  \hspace{1mm}
  \begin{subfigure}[b]{0.45\textwidth}
    \centering
    \includegraphics[width=\textwidth]{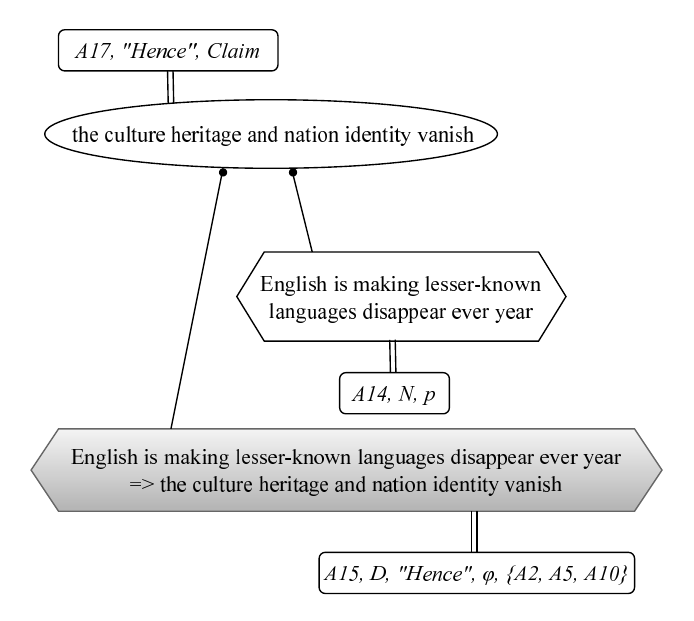}
    \caption{}
    \label{fig:image7}
  \end{subfigure}
  \caption{(a) In a portion of an AKG created from \textit{essay056} of the \textit{AAEC version2} dataset \cite{stab2017parsing}, \textit{conclusion} $A_{17}$ has both incoming \textit{modus ponens} (indirect relation) and \textit{support} (direct relation) edges. (b) The \textit{support} edge is discarded in the proposed AKG.}
  \label{fig:combined2}
\end{figure}

\noindent  \textit{premises}, and $a$ denotes \textit{assumptions}. For instance, a possible set of attributes for a \textit{premise} of type \textit{ordinary} can be $\{``however", p\}$. The set of attributes for \textit{implicit premises} is the same as that for \textit{premises} (explicit ones). We maintain markers contained in \textit{premises} and \textit{implicit premises} as attributes to serve as auxiliary information. \textit{Agreement} and \textit{contrary} edges have no attributes. 

After constructing the KB graph, we derive the argument set from it. We infer \textit{conclusions} from the KB graph by applying \textit{modus ponens} to an \textit{inference rule premise} and its antecedent \textit{premise}. All \textit{inference rule premises}, \textit{premises}, \textit{implicit premises}, \textit{conclusions}, and \textit{implicit conclusions} together form the argument set.

\subsection{Construction of an AKG}
In this phase, we construct the AKG from the KB graph and the argument set. Each argument in an AKG is assigned a unique argument ID as an attribute. All the components of an AKG are illustrated in Figure \ref{fig:image4}. Table \ref{tab:table1} presents a list of attributes and their example values in the AKG.

In the case of \textit{inference rule premises}, the rule IDs in the KB graph are replaced with argument IDs in the corresponding AKG. A potential set of attributes for a defeasible \textit{inference rule premise} in an AKG can be represented as $\{A_1, D, ``therefore", \{A_2, A_3\}, \phi\}$, where $A_1$, $A_2$, and $A_3$ are the argument IDs of the respective \textit{inference rule premises}, and the order of the attributes is preserved as described in Section 3.1.

Likewise, the set of attributes for a \textit{premise} and an \textit{implicit premise} in an AKG is the same as in the corresponding KB graph, except that an additional attribute for the argument ID is included at the beginning of the attribute set. For instance, a possible set of attributes for a \textit{premise} of type \textit{ordinary} in an AKG can be $\{A_4, ``however", p\}$, where $A_4$ is its argument ID.

\textit{Conclusion} nodes are newly introduced in the AKG. \textit{Conclusions} can either be inferred by the rule of modus ponens or be explicitly annotated in a dataset. The attributes of a \textit{conclusion} node include its argument ID and any marker attached to the \textit{conclusion}. In the absence of a marker, the corresponding field is filled with $N$. For instance, a possible set of attributes for a \textit{conclusion} node can be $\{A_5, N\}$, where $A_5$ is its argument ID and $N$ denotes that no marker is attached to that \textit{conclusion}. The set of attributes for \textit{implicit conclusions} is the same as that for \textit{conclusions} (explicit ones). We retain the markers contained in \textit{conclusions} and \textit{implicit conclusions} as attributes, serving as auxiliary information just as in \textit{premises} and \textit{implicit premises}.

Lastly, \textit{contrary} and \textit{agreement} edges in the KB graph are transformed into \textit{attack} and \textit{support} edges, respectively, in the AKG to reflect conflicts and agreements between arguments. An \textit{attack} edge has one attribute: the type of attack, as mentioned in Section 2. \textit{Reb} denotes a rebuttal attack, \textit{UM} denotes an undermining attack, and \textit{UC} denotes an undercutting attack. \textit{Modus ponens} edges are new additions in the AKG. Their main purpose is to connect an \textit{inference rule premise} and its corresponding antecedent \textit{premise} to a \textit{conclusion}. However, sometimes a \textit{modus ponens} edge may connect an \textit{inference rule premise} and its antecedent \textit{premise} to another \textit{premise}, which is essentially an intermediate \textit{conclusion} by nature but either functions as a \textit{premise} for another \textit{conclusion} in the chain of reasoning or is annotated as a \textit{premise} in the dataset. We retain them as \textit{premises} in the AKG. The scenario is illustrated in Figure \ref{fig:image5}. \textit{Modus ponens} edges and \textit{support} edges have no attributes.

In some cases, the original dataset contains a \textit{support} edge (direct relation) from one \textit{premise} to another \textit{premise} or a \textit{conclusion}. However, there may also be \textit{modus ponens} edges (indirect relation) connecting an \textit{inference rule premise} and the first \textit{premise} to the second \textit{premise} or the \textit{conclusion}. In such cases, we discard the \textit{support} edge and retain only the \textit{modus ponens} edges, as our goal is to use this AKG to help reasoning models \cite{kambhampati2024can, yu2024natural} to learn the implicit indirect relations. Figure \ref{fig:combined2} illustrates this scenario.

\subsection{Detection of inference markers (IM) indicative of inference rules}
We use markers to identify inference rules, which are not annotated in existing datasets. 
Markers are linguistic expressions that signal relationships between different parts of a text or conversation. They may consist of a single word, a phrase, or an entire sentence \cite{tseronis2011towards}. Markers help to identify support and attack relations between argumentative components \cite{lawrence2015combining}. Some markers typically introduce a premise or reason, while others
\newpage
\begin{table}[H] 
\centering
\small
\caption{A non-exhaustive list of inference markers (IMs) and their indicator types, extracted from the AAEC Version2 \cite{stab2017parsing} dataset.}
\label{tab:table2}
\begin{tabular}{ll}
\toprule
\textbf{Inference Marker (IM)} & \textbf{Indicator Type (Stab \& Gurevych \cite{stab2017parsing})} \\
\midrule
\texttt{because}            & Premise \\
\texttt{since}              & Premise \\
\texttt{given that}         & Premise \\
\texttt{due to}             & Premise \\
\texttt{in view of}         & Premise \\
\texttt{in light of}         & Premise \\
\texttt{for the reason that}         & Premise \\
\texttt{as}         & Premise \\
\texttt{deduced}         & Premise \\
\texttt{derived from}         & Premise \\
\texttt{may be inferrred}         & Premise \\
\texttt{in that}         & Premise \\
\texttt{so}                 & Claim \\
\texttt{therefore}          & Claim \\
\texttt{thus}               & Claim \\
\texttt{hence}              & Claim \\
\texttt{as a result}        & Claim \\
\texttt{it follows that}    & Claim \\
\texttt{follows that}                 & Claim \\
\texttt{we may deduce}      & Claim \\
\texttt{implies}            & Claim \\
\texttt{accordingly}                 & Claim \\
\texttt{consequently}                 & Claim \\
\texttt{conclude that}                 & Claim \\
\texttt{entails}                 & Claim \\
\texttt{proves that}                 & Claim \\
\texttt{shows that}                 & Claim \\
\texttt{suggests that}                 & Claim \\
\bottomrule
\end{tabular}
\end{table}

\noindent introduce conclusions or inferred results \cite{stab2014identifying}. Thus, they relate to inference either by presenting a reason or a result.

\textit{Reasoning markers} (RM) \cite{clayton2022predicting} form a proper subset of general discourse markers (DM), used to express logical links between claims and premises. Kuribayashi et al. \cite{kuribayashi2019empirical} used the term \textit{argumentative markers} (AM) to refer to conjunctive expressions that are frequently used in argumentative texts. Stab and Gurevych \cite{stab2014identifying} further distinguished such AMs from general DMs, such as those indicating temporal relations. In this work, we focus on identifying a specific set of markers that signal the presence of inference rules within discourse. We refer to this set as \textit{inference markers} (IM). We observe certain patterns and propose the following heuristic rules for detecting IMs.

\begin{enumerate}
\item \textbf{Forward IM at the beginning of a sentence:} If a sentence begins with a marker followed by a comma, it likely expresses a conclusion drawn from the previous sentence(s), indicating an inference rule from antecedent to consequent. Potential examples of such IMs include \textit{as a result}, \textit{so}, \textit{hence}, \textit{thus}, \textit{therefore} etc. \textit{A} denotes \textit{antecedent} and \textit{C} denotes \textit{consequent} in all the following examples.\\ \\
\textbf{Example:} \([ \text{She was the most experienced candidate.} ]_{\text{A}} \quad [ \text{Therefore,} ]_{\text{IM}}\\ \quad [ \text{She was selected for the position.} ]_{\text{C}} \)\\ 

\item \textbf{Forward IM between sentences or clauses:} If a marker appears between two complete sentences or clauses (especially after a semicolon or period), then it likely connects a premise (antecedent) to a claim (consequent), indicating an inference. Such IMs include \textit{as a result}, \textit{so}, \textit{hence}, \textit{thus}, \textit{therefore} etc.\\ \\
\textbf{Example:} \([ \text{The evidence was overwhelming;} ]_{\text{A}} \quad [ \text{thus,} ]_{\text{IM}} \quad [ \text{the jury returned a guilty verdict.} ]_{\text{C}} \)\\

\item \textbf{Backward IM linking cause to effect:}
If a causal marker links two clauses, then the sentence expresses a causal inference, where the second clause serves as the premise (antecedent) and the first clause represents the conclusion (consequent). Such IMs include \textit{due to the fact that}, \textit{seeing that}, \textit{because}, \textit{as}, \textit{since} etc.\\ \\
\textbf{Example:} \([ \text{The event was canceled } ]_{\text{C}} \quad [ \text{due to the fact that } ]_{\text{IM}} \quad [ \text{there was a storm.} ]_{\text{A}} \)
\end{enumerate}

\subsubsection{Detection of implicit inference rules}
Argumentative relations frequently remain implicit, without being signaled by DMs \cite{stab2014identifying}. In cases of implicit inference, end-of-sentence punctuation from the previous sentence can be treated as an IM. A similar idea for AM extraction was proposed by Kuribayashi et al. \cite{kuribayashi2019empirical}.

In the AKReF, we explicitly encode inference rules and \textit{modus ponens} edges in the proposed AKG. This AKG format is expected to enable the reasoning models to learn latent inference rules without relying on explicit IMs.

\subsubsection{Manually compiled IMs from the AAEC Version2 dataset}
In this paper, our main focus is on proposing an argumentative knowledge representation framework. We do not aim to develop a nearly exhaustive list of IMs. However, Stab and Gurevych \cite{stab2017parsing} compiled two lists of claim and premise indicators from the AAEC Version 2 dataset. We extract potential IMs from these lists and present them in Table  \ref{tab:table2}.

\section{Case study}
\subsection{An example from the \textit{AAEC Version2} dataset}
\begin{figure}[H]
    \centering
    \includegraphics[width=0.95\linewidth, height=0.37\textheight, keepaspectratio]{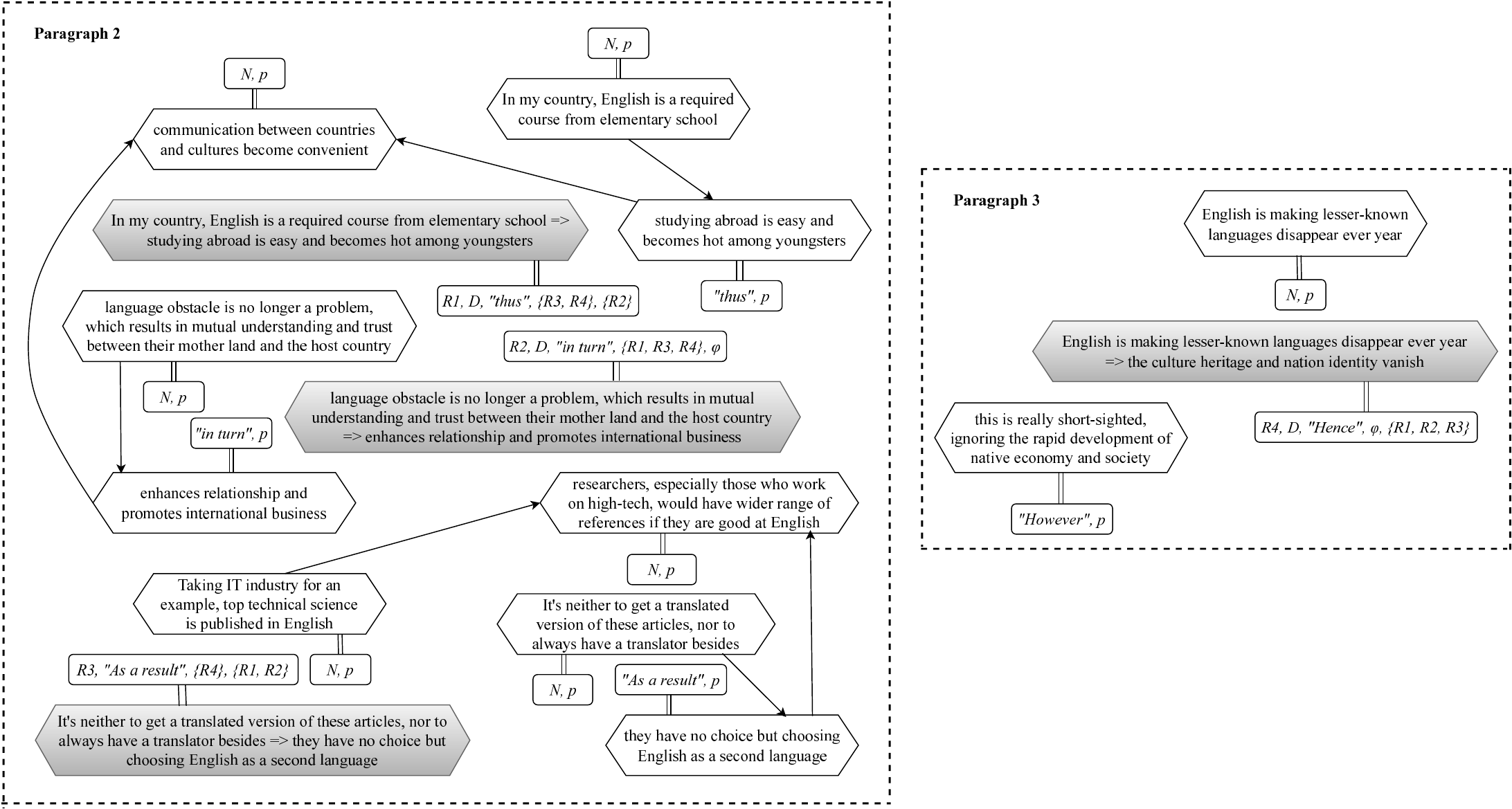}
    \caption{The KB graph for \textit{essay056} from \textit{AAEC version 2} dataset \cite{stab2017parsing}. A preference order of $A_5 > A_2 > A_{10} > A_{15}$ is imposed among the defeasible inference rules. Refer to Figure \ref{fig:image4} for symbol details.}
    \label{fig:image8}
\end{figure}

We construct a KB and subsequently the KB graph from \textit{essay056} of the \textit{AAEC Version2} dataset \cite{stab2017parsing}. We impose a preference order of $A_5 > A_2 > A_{10} > A_{15}$ over the defeasible inference rules. The KB is presented in the appendix. The KB graph corresponding to the KB is shown in Figure  \ref{fig:image8}. 

Next, we construct the arguments from the KB. For each argument, we indicate its type as P, IRP, or C, where these stand for \textit{premise}, \textit{inference rule premise}, and \textit{conclusion}, respectively. In some intermediate cases, although an \textit{inference rule premise} and its antecedent \textit{premise} lead to a consequent by applying \textit{modus ponens}, we mark the 

\pagebreak

\vspace{2mm}

\begin{figure}[H]
    \centering
    \includegraphics[width=0.95\linewidth, height=0.45\textheight, keepaspectratio]{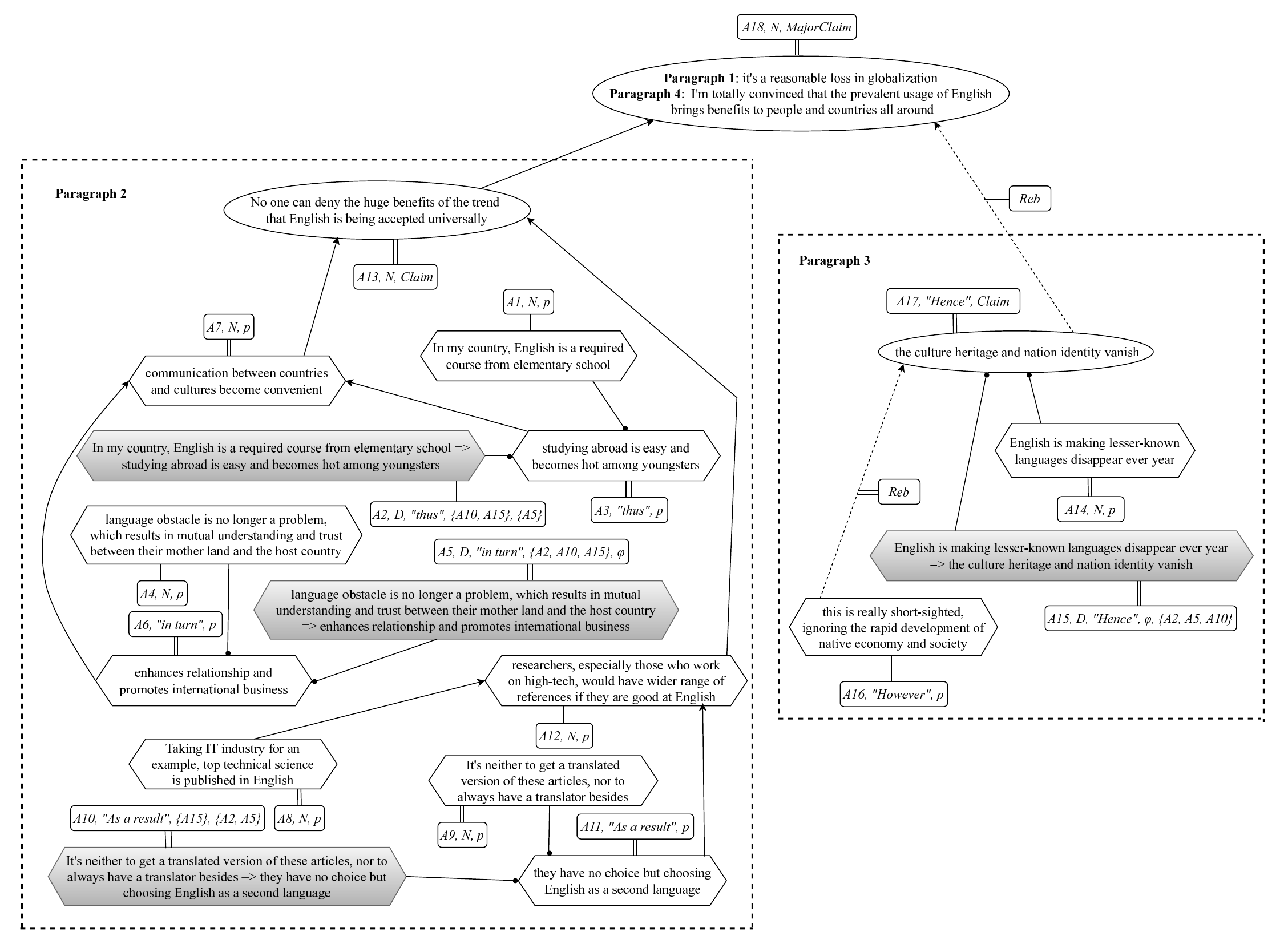}
    \caption{The AKG corresponding to the KB graph in Figure \ref{fig:image8}, originally created from \textit{essay056} of the \textit{AAEC version 2} dataset \cite{stab2017parsing}. Refer to Figure \ref{fig:image4} for symbol details.}
    \label{fig:image9}
\end{figure}
\pagebreak

\noindent consequent as a \textit{premise} rather than a \textit{conclusion}, as explained in Section 3.2 and Figure \ref{fig:image5}. We include these \textit{premises} in the KB, the KB graph, and eventually in the AKG. We include the annotated \textit{claims} and \textit{major claims} in the dataset as \textit{conclusions} in the argument set. The corresponding argument set and the interactions between them are presented in the appendix.

Figure \ref{fig:image9} shows the AKG corresponding to the argument set and its interactions as provided in the appendix. For this dataset, we add one new attribute to the \textit{conclusion} nodes in addition to those mentioned in Section 3.2. This newly added attribute has two possible types: \textit{Claim} and \textit{MajorClaim}. These are added because the \textit{conclusion} in our argument set was originally labeled as a \textit{claim} or \textit{major claim} in the dataset, providing auxiliary information. Each \textit{claim} in the dataset has a \textit{stance} attribute indicating whether it is \textit{for} or \textit{against} the \textit{MajorClaim} in the original dataset. We convert each \textit{for stance} into a \textit{support} edge and each \textit{against stance} into an \textit{attack} edge. 

From the interactions between the arguments presented in the appendix, we observe that $A_3$ has incoming \textit{modus ponens} edges from $A_1$ and $A_2$, in addition to a \textit{support} edge from $A_1$. Therefore, the \textit{modus ponens} edges are retained, and the \textit{support} edge is discarded, as mentioned in Section 3.2 and Figure \ref{fig:combined2}. Similar cases occur with $A_6$, $A_{11}$, and $A_{17}$. In all these cases, we retain the \textit{modus ponens} edges and discard the \textit{support} edges. In this manner, we applied our proposed KB graph and AKG to this dataset with minor adjustments.

\section{An application of the proposed AKG representation}
The proposed AKG visually represents the ACs, ARs, and metadata attributes in a clear and comprehensible manner. This graphical knowledge representation facilitates further complex analyses to assess the acceptability of argument sets. To achieve this, we compute the Naïve and Preferred semantics from the AKG in Figure \ref{fig:image9}.

Let $S_{para2}$ be the set of arguments derived from paragraph 2 of the essay, as shown in Figure \ref{fig:image9}. Hence,

$S_{para2} = \{A_1, A_2, A_3, A_4, A_5, A_6, A_7, A_8, A_9, A_{10}, A_{11}, A_{12}, A_{13}\}$

\noindent Likewise, let $S_{para3}$ be the set of arguments derived from paragraph 3, and $S_{essay}$ be the set of arguments derived from the entire essay. Hence,

$S_{para3} = \{A_{14}, A_{15}, A_{16}, A_{17}\}$\\
\indent $S_{essay} = \{A_1, A_2, A_3, A_4, A_5, A_6, A_7, A_8, A_9, A_{10}, A_{11}, A_{12}, A_{13}, A_{14}, A_{15}, A_{16}, A_{17}, A_{18}\}$

\noindent From the AKG, it is evident that $S_{para2}$ is both conflict-free and admissible, because none of the arguments in $S_{para2}$ attacks any other in the set, and no argument external to $S_{para2}$ attacks any argument within it.  On the other hand, $S_{para3}$ is not conflict-free because argument $A_{16}$ (\textit{this is really short-sighted, ignoring the rapid development of native economy and society}) within it attacks argument $A_{17}$ (\textit{the culture heritage and nation identity vanish}), which is also in $S_{para3}$. Therefore, $S_{para3}$ is not admissible. In the same manner, $S_{essay}$ is not conflict-free ($A_{16}$ attacks $A_{17}$ and $A_{17}$ attacks $A_{18}$), and, therefore, not admissible. $S_{para2} \bigcup A_{18}$ is conflict-free because there is no attack relation between the arguments in the set. The set is not admissible because argument $A_{18}$ (\textit{Paragraph 1: it's a reasonable loss in globalization. Paragraph 4: I'm totally convinced that the prevalent usage of English brings benefits to people and countries all around.}) in the set is attacked by the external argument $A_{17}$ (\textit{the culture heritage and nation identity vanish}), and no argument from the set successfully attacks $A_{17}$.

Next, the Naïve semantics from this AKG is the following set, as it is the maximal conflict-free set that cannot be further expanded while remaining conflict-free.

$S_{NS} = \{A_1, A_2, A_3, A_4, A_5, A_6, A_7, A_8, A_9, A_{10}, A_{11}, A_{12}, A_{13}, A_{14}, A_{15}, A_{16}, A_{18}\}$, where $S_{NS}$ represents the Naïve semantics derived from the AKG.

\indent In this case, $S_{NS} = S_{PS}$, where $S_{PS}$ represents the Preferred semantics derived from the AKG because it is the maximal (w.r.t. set inclusion) admissible set too. Argument $A_{18}$ (\textit{Paragraph 1: it's a reasonable loss in globalization. Paragraph 4: I'm totally convinced that the prevalent usage of English brings benefits to people and countries all around.}) in the Preferred semantics is attacked by the external argument $A_{17}$ (\textit{the culture heritage and nation identity vanish}), which is attacked by the internal argument $A_{16}$ (\textit{this is really short-sighted, ignoring the rapid development of native economy and society}). 

\section{Related work}
Argument diagramming \cite{reed2007argument} is a foundational technique for visualizing and analyzing argument structures. Originating in the field of informal logic, it has been widely applied in law, artificial intelligence, and critical thinking \cite{reed2007argument}. The primary purpose of these diagrams is to reveal the logical structure of arguments by identifying premises, conclusions, and the inferential connections between them.

Early uses of argument diagrams appeared in 19\textsuperscript{th} and 20\textsuperscript{th} century logic textbooks. Richard Whately was among the first to represent arguments diagrammatically, introducing the notion of a “chain of arguments” by starting with the conclusion and tracing back to its supporting reasons \cite{reed2007argument}. Beardsley enhanced this approach by introducing circled numbers to represent statements and arrows to connect premises to conclusions. Further, that work categorized arguments as convergent, divergent, or serial, and used diagrams to illustrate logical fallacies such as circular reasoning \cite{reed2007argument}.

A significant shift in argumentation theory was introduced by Toulmin in \textit{The Uses of Argument}. His model added components such as the warrant, which connects data to the claim, the Qualifier to express the strength of inference, the rebuttal to challenge conclusions, and backing to strengthen the warrant  \cite{reed2007argument}. Freeman further refined argument diagramming by distinguishing between linked and convergent arguments and introducing the concept of supposition and modality of the argument  \cite{reed2007argument}. Pollock (2002) expanded Toulmin’s rebuttal by formalizing the concepts of rebutting defeaters, which attack the conclusion directly, and undercutting defeaters, which challenge the inferential connection between premise and conclusion  \cite{reed2007argument}.

In recent computational approaches, Al-Khatib et al. \cite{al2020end} proposed a model for constructing an AKG and introduced a new argumentation corpus. In their framework, nodes represent concept instances, while directed edges denote unweighted effect relations. Concept consequences and groundings are captured as node attributes. Building on this, subsequent work \cite{al2021employing} constructed three types of graphs: (1) a ground-truth knowledge graph (KG) manually annotated based on the corpus from Al-Khatib et al. \cite{al2020end}; (2) a generated KG automatically derived using the argument knowledge relation extraction approach \cite{al2020end}; and (3) a causality KG focusing specifically on causal relations between the concepts. Unlike these works, our proposed method constructs an AKG using pre-annotated ACs and ARs. Moreover, we add metadata as attributes and use markers to identify inference rules. We also annotate three attack types and add \textit{modus ponens} edges to capture indirect relations. These enhancements support learning indirect relations and reasoning.

\section{Conclusion}
In this work, we use our proposed framework, AKReF, to enhance the basic dataset annotations by constructing a structured KB graph enriched with metadata encoded as node attributes. From this graph, arguments are systematically derived using premises, inference rules, and modus ponens. These arguments form the basis of an AKG, where both nodes and edges carry informative attributes. To address the absence of inference rules in existing annotations, we identify them by locating IMs. We incorporate the inference rules as a new type of premise node in both the KB and the AKG. This addition allows for detecting undercut attacks that were not identifiable in the original annotations. Our proposed KB graph and AKG present the argumentative structure of the texts in a graphical manner. It is relatively easier to comprehend than the textual format, where domain knowledge is mandatory. In the future, existing graph neural network models can be leveraged to exploit the inference engine and learn the argumentative structures from the resulting AKG. Argumentative relations are often implicit and not indicated by markers \cite{stab2014identifying}. Our proposed method could lay the foundation for reasoning models \cite{kambhampati2024can, yu2024natural} to learn indirect relations, which may subsequently contribute to complex argumentative tasks such as belief revision \cite{snaith2017argument} and argument evaluation \cite{baroni2015automatic, wachsmuth2017argumentation}.

\section*{Appendix}
\subsection*{The KB constructed from \textit{essay056} in the \textit{AAEC Version2} Dataset \cite{stab2017parsing}}
All the notations related to a KB and an EKB are explained in Section 2 and Section 3 respectively.\\ 
\noindent
$R_s = \phi$;\\
\\
$
R_d = \{\\
\textit{In my country, English is a required course from elementary school} \Rightarrow \\
\textit{studying abroad is easy and becomes hot among youngsters}; \\
\\
\textit{language obstacle is no longer a problem, which results in mutual}\\
\textit{understanding and trust between their mother land and the host country}
\Rightarrow \\ \textit{enhances relationship and promotes international business}; \\
\\
\textit{It's neither to get a translated version of these articles, nor to
always have}\\ \textit{a translator besides} \Rightarrow \textit{they have no choice but
choosing English as a second}\\ \textit{language};\\
\\
\textit{English is making lesser-known languages disappear ever year}\Rightarrow \textit{the culture}\\ \textit{heritage and nation identity vanish};\\
\}
$
\\
\vspace{10pt} 
\\
$
\noindent K = \{\\
\textit{In my country, English is a required course from elementary school}; \\
\\
\textit{In my country, English is a required course from elementary school} \Rightarrow \\
\textit{studying abroad is easy and becomes hot among youngsters}; \\
\\
\textit{studying abroad is easy and becomes hot among youngsters};\\
\\
\textit{language obstacle is no longer a problem, which results in mutual}\\
\textit{understanding and trust between their mother land and the host country};\\
\\
\textit{language obstacle is no longer a problem, which results in mutual}\\
\textit{understanding and trust between their mother land and the host country}
\Rightarrow \\ \textit{enhances relationship and promotes international business}; \\
\\
\textit{enhances relationship and promotes international business};\\
\\
\textit{communication between countries and cultures become convenient};\\
\\
\textit{Taking IT industry for an example, top technical science is published in}\\
\textit{English;}\\
\\
\textit{It's neither to get a translated version of these articles, nor to always have}\\
\textit{a translator besides};\\
\\
\textit{It's neither to get a translated version of these articles, nor to
always have}\\ \textit{a translator besides} \Rightarrow \textit{they have no choice but
choosing English as a second}\\ \textit{language};\\
\\
\textit{they have no choice but choosing English as a second language};\\
\\
\textit{researchers, especially those who work on high-tech, would have wider range}\\
\textit{of references if they are good at English};\\
\\
\textit{English is making lesser-known languages disappear ever year};\\
\\
\textit{English is making lesser-known languages disappear ever year}\Rightarrow \textit{the culture}\\ \textit{heritage and nation identity vanish};\\
\\
\textit{this is really short-sighted, ignoring the rapid development of native economy}\\
\textit{and society};\\
\}
$
\\
\vspace{10pt} 
\\
$C(KB) = \phi$;\\
\vspace{10pt} 
\\
$
Ag(KB) = \{\\
\textit{In my country, English is a required course from elementary school}\\
\in \underline{\textit{studying abroad is easy and becomes hot among youngsters}};\\
\\
\textit{studying abroad is easy and becomes hot among youngsters}\in\\
\underline{\textit{communication between countries and cultures become convenient}};\\
\\
\textit{language obstacle is no longer a problem, which results in mutual}\\
\textit{understanding and trust between their mother land and the host}\\
\textit{country} \in \underline{\textit{enhances relationship and promotes international business}};\\
\\
\textit{enhances relationship and promotes international business} \in \\ 
\underline{\textit{communication between countries and cultures become convenient}};\\ 
\\
\textit{Taking IT industry for an example, top technical science is published}\\
\textit{in English} \in \underline{\textit{researchers, especially those who work on high-tech,}}\\ 
\underline{\textit{would have wider range of references if they are good at English}};\\
\\
\textit{It's neither to get a translated version of these articles, nor to always}\\
\textit{have a translator besides} \in \underline{\textit{they have no choice but choosing English }}\\
\underline{\textit{as a second language}};\\
\\
\textit{they have no choice but choosing English} \in \underline{\textit{researchers, especially those}}\\
\underline{\textit{who work on high-tech, would have wider range of references if they are}}\\ \underline{\textit{good at English}};\\
\}
$ 

\subsection*{Argument set corresponding to the above KB}
\noindent $A_1: \textit{In my country, English is a required course from elementary school}$; (P)\\
\\
$A_2: \textit{In my country, English is a required course from elementary school} \Rightarrow \\
\textit{studying abroad is easy and becomes hot among youngsters}$; (IRP)\\
\\
$A_3: A_2, A_1 \vdash \textit{studying abroad is easy and becomes hot among youngsters}$; (P)\\
\\
$A_4: \textit{language obstacle is no longer a problem, which results in mutual}\\
\textit{understanding and trust between their mother land and the host country}$; (P)\\
\\
$A_5: \textit{language obstacle is no longer a problem, which results in mutual}\\
\textit{understanding and trust between their mother land and the host country}
\Rightarrow \\ \textit{enhances relationship and promotes international business}$; (IRP)\\
\\
$A_6: A_5, A_4 \vdash \textit{enhances relationship and promotes international business}$; (P)\\
\\
$A_7: \textit{communication between countries and cultures become convenient}$; (P)\\
\\
$A_8: \textit{Taking IT industry for an example, top technical science is published in English}$; (P)\\
\\
$A_9: \textit{It's neither to get a translated version of these articles, nor to always have a}\\ 
\textit{translator besides}$; (P)\\
\\
$A_{10}: \textit{It's neither to get a translated version of these articles, nor to
always have}\\ \textit{a translator besides} \Rightarrow \textit{they have no choice but
choosing English as a second}\\ \textit{language}$; (IRP)\\
\\
$A_{11}: A_{10},  A_9 \vdash \textit{they have no choice but choosing English as a second language}$; (P)\\
\\
$A_{12}: \textit{researchers, especially those who work on high-tech, would have wider range}\\ \textit{of references if they are good at English}$; (P)\\
\\
$A_{13}: \textit{No one can deny the huge benefits of the trend that English is being accepted}\\ \textit{universally}$; (C)\\
\\
$A_{14}: \textit{English is making lesser-known languages disappear ever year}$; (P)\\
\\
$A_{15}: \textit{English is making lesser-known languages disappear ever year}\Rightarrow \textit{the culture}\\ \textit{heritage and nation identity vanish}$; (IRP)\\
\\
$A_{16}: \textit{this is really short-sighted, ignoring the rapid development of native}\\ 
\textit{economy and society}$; (P)\\
\\
$A_{17}: A_{15}, A_{14} \vdash \textit{the culture heritage and nation identity vanish}$; (C)\\
\\
$A_{18}: \textit{It's a reasonable loss in globalization};\\
\textit{I'm totally convinced that the prevalent usage of English brings benefits to people}\\
\textit{and countries all around}$; (C)

\subsection*{Interactions among the arguments in the argument set}
The \textit{support} and \textit{attack} relations between the arguments are as follows.\\
\\
$A_1$ supports $A_3$; $A_3$ supports $A_7$; $A_4$ supports $A_6$; $A_6$ supports $A_7$; $A_8$ supports $A_{12}$; $A_9$ supports $A_{11}$; $A_{11}$ supports $A_{12}$; $A_{12}$ supports $A_{13}$; $A_{14}$ supports $A_{17}$; $A_{16}$ attacks $A_{17}$; $A_{13}$ supports $A_{18}$; $A_{17}$ attacks $A_{18}$.\\
\\
The following are the \textit{modus ponens} relations between the arguments.\\
\\
$A_1$ and $A_2$ infer $A_3$ via \textit{modus ponens}; $A_4$ and $A_5$ infer $A_6$ via \textit{modus ponens}; $A_9$ and $A_{10}$ infer $A_{11}$ via \textit{modus ponens}; $A_{14}$ and $A_{15}$ infer $A_{17}$ via \textit{modus ponens}.

\bibliographystyle{plain}
\bibliography{references} 

\begin{thebibliography}{10}

\bibitem{al2020end}
Khalid Al-Khatib, Yufang Hou, Henning Wachsmuth, Charles Jochim, Francesca
  Bonin, and Benno Stein.
\newblock End-to-end argumentation knowledge graph construction.
\newblock In {\em Proceedings of the AAAI conference on artificial
  intelligence}, volume~34, pages 7367--7374, 2020.

\bibitem{al2021employing}
Khalid Al~Khatib, Lukas Trautner, Henning Wachsmuth, Yufang Hou, and Benno
  Stein.
\newblock Employing argumentation knowledge graphs for neural argument
  generation.
\newblock In {\em Proceedings of the 59th Annual Meeting of the Association for
  Computational Linguistics and the 11th International Joint Conference on
  Natural Language Processing (Volume 1: Long Papers)}, pages 4744--4754, 2021.

\bibitem{amgoud2006final}
Leila Amgoud, Lianne Bodenstaff, Martin Caminada, S~McBurney, H~Prakken,
  J~Veenen, and GAW Vreeswijk.
\newblock Final review and report on formal argumentation system.
\newblock 2006.

\bibitem{baroni2015automatic}
Pietro Baroni, Marco Romano, Francesca Toni, Marco Aurisicchio, and Giorgio
  Bertanza.
\newblock Automatic evaluation of design alternatives with quantitative
  argumentation.
\newblock {\em Argument \& Computation}, 6(1):24--49, 2015.

\bibitem{carstens2015towards}
Lucas Carstens and Francesca Toni.
\newblock Towards relation based argumentation mining.
\newblock In {\em Proceedings of the 2nd Workshop on Argumentation Mining},
  pages 29--34, 2015.

\bibitem{clayton2022predicting}
Jonathan Clayton and Robert Gaizauskas.
\newblock Predicting the presence of reasoning markers in argumentative text.
\newblock In {\em Proceedings of the 9th Workshop on Argument Mining}, pages
  137--142, 2022.

\bibitem{dung1995acceptability}
Phan~Minh Dung.
\newblock On the acceptability of arguments and its fundamental role in
  nonmonotonic reasoning, logic programming and n-person games.
\newblock {\em Artificial intelligence}, 77(2):321--357, 1995.

\bibitem{feng2011classifying}
Vanessa~Wei Feng and Graeme Hirst.
\newblock Classifying arguments by scheme.
\newblock In {\em Proceedings of the 49th annual meeting of the association for
  computational linguistics: Human language technologies}, pages 987--996,
  2011.

\bibitem{kambhampati2024can}
Subbarao Kambhampati.
\newblock Can large language models reason and plan?
\newblock {\em Annals of the New York Academy of Sciences}, 1534(1):15--18,
  2024.

\bibitem{kuribayashi2019empirical}
Tatsuki Kuribayashi, Hiroki Ouchi, Naoya Inoue, Paul Reisert, Toshinori
  Miyoshi, Jun Suzuki, and Kentaro Inui.
\newblock An empirical study of span representations in argumentation structure
  parsing.
\newblock In {\em Proceedings of the 57th annual meeting of the association for
  computational linguistics}, pages 4691--4698, 2019.

\bibitem{lawrence2015combining}
John Lawrence and Chris Reed.
\newblock Combining argument mining techniques.
\newblock In {\em Proceedings of the 2nd Workshop on Argumentation Mining},
  pages 127--136, 2015.

\bibitem{lawrence2020argument}
John Lawrence and Chris Reed.
\newblock Argument mining: A survey.
\newblock {\em Computational Linguistics}, 45(4):765--818, 2020.

\bibitem{pollock1987defeasible}
John~L Pollock.
\newblock Defeasible reasoning.
\newblock {\em Cognitive science}, 11(4):481--518, 1987.

\bibitem{prakken2010abstract}
Henry Prakken.
\newblock An abstract framework for argumentation with structured arguments.
\newblock {\em Argument \& Computation}, 1(2):93--124, 2010.

\bibitem{reed2007argument}
Chris Reed, Douglas Walton, and Fabrizio Macagno.
\newblock Argument diagramming in logic, law and artificial intelligence.
\newblock {\em The Knowledge Engineering Review}, 22(1):87--109, 2007.

\bibitem{ruiz2022automatic}
Ramon Ruiz-Dolz, Stella Heras, and Ana Garc{\'\i}a-Fornes.
\newblock Automatic debate evaluation with argumentation semantics and natural
  language argument graph networks.
\newblock {\em arXiv preprint arXiv:2203.14647}, 2022.

\bibitem{snaith2017argument}
Mark Snaith and Chris Reed.
\newblock Argument revision.
\newblock {\em Journal of Logic and Computation}, 27(7):2089--2134, 2017.

\bibitem{stab2014identifying}
Christian Stab and Iryna Gurevych.
\newblock Identifying argumentative discourse structures in persuasive essays.
\newblock In {\em Proceedings of the 2014 conference on empirical methods in
  natural language processing (EMNLP)}, pages 46--56, 2014.

\bibitem{stab2017parsing}
Christian Stab and Iryna Gurevych.
\newblock Parsing argumentation structures in persuasive essays.
\newblock {\em Computational Linguistics}, 43(3):619--659, 2017.

\bibitem{tseronis2011towards}
A~Tseronis et~al.
\newblock Towards an empirically plausible classification of argumentative
  markers.
\newblock 2011.

\bibitem{wachsmuth2017argumentation}
Henning Wachsmuth, Nona Naderi, Ivan Habernal, Yufang Hou, Graeme Hirst, Iryna
  Gurevych, and Benno Stein.
\newblock Argumentation quality assessment: Theory vs. practice.
\newblock In {\em Proceedings of the 55th Annual Meeting of the Association for
  Computational Linguistics (Volume 2: Short Papers)}, pages 250--255, 2017.

\bibitem{yu2024natural}
Fei Yu, Hongbo Zhang, Prayag Tiwari, and Benyou Wang.
\newblock Natural language reasoning, a survey.
\newblock {\em ACM Computing Surveys}, 56(12):1--39, 2024.

\end{thebibliography}

\end{document}